\documentclass[runningheads,dvipsnames]{llncs}
\usepackage{graphicx}
\usepackage{booktabs}
\usepackage{bm}
\usepackage{amsfonts}
\usepackage{amsmath}
\usepackage{paralist}
\usepackage{multirow}
\usepackage{tabularx}
\usepackage{tikz}
\usepackage{algorithm}
\usepackage{algpseudocode}
\usepackage{xspace}
\usepackage{capt-of}
\usepackage[]{graphicx}
\usepackage{hyperref}

\usepackage{array}
\usepackage{amsmath}
\DeclareMathOperator*{\argmax}{arg\,max}

\usepackage[textsize=small
            ]{todonotes}

\newcolumntype{L}[1]{>{\raggedright\let\newline\\\arraybackslash\hspace{0pt}}m{#1}}
\newcolumntype{C}[1]{>{\centering\let\newline\\\arraybackslash\hspace{0pt}}m{#1}}
\newcolumntype{R}[1]{>{\raggedleft\let\newline\\\arraybackslash\hspace{0pt}}m{#1}}

\hypersetup{
	pdftoolbar=true,        
	pdfmenubar=true,        
	pdffitwindow=false,     
	pdfstartview={FitH},    
	pdftitle={},    
	pdfauthor={},     
	pdfsubject={},   
	pdfcreator={},   
	pdfproducer={}, 
	pdfkeywords={}, 
	pdfnewwindow=true,      
	colorlinks=true,       
	  linkcolor=Brown,          
	  citecolor=OliveGreen,        
	  filecolor=magenta,      
	  urlcolor=NavyBlue           
}

\begin{document}
\title{Knowledge-Driven Modulation of Neural Networks with Attention Mechanism for Next Activity Prediction
}
\titlerunning{Knowledge-Driven Modulation of Neural Networks}
%
\author{Ivan Donadello$^\ast$ \and Jonghyeon Ko$^\ast$ \and
Fabrizio Maria Maggi$^\ast$ \and Jan Mendling$^\dagger$ \and
Francesco Riva$^\ddagger$ \and Matthias Weidlich$^\dagger$}
\authorrunning{I. Donadello et al.}
%
\institute{Free University of Bozen-Bolzano, Bolzano, Italy
\email{\{Ivan.Donadello,lncs\}@uni-heidelberg.de}\\
\url{http://www.springer.com/gp/computer-science/lncs} \and
ABC Institute, Rupert-Karls-University Heidelberg, Heidelberg, Germany\\
\email{lncs@springer.com}
}
%
%

\institute{$^\ast$Free University of Bozen-Bolzano, Bolzano, Italy \\
\email
\{ivan.donadello, jongko\}@unibz.it, maggi@inf.unibz.it \\
$^\dagger$Humboldt-Universität zu Berlin, Berlin, German \\
\email
\{jan.mendling, matthias.weidlich\}@hu-berlin.de \\
$^\ddagger$Datalane SRL, Verona, Italy \\
\email
f.riva@datalane.nl \\
}

\maketitle              
\begin{abstract}
Predictive Process Monitoring (PPM) aims at leveraging historic process execution data to predict how ongoing executions will continue up to their completion. In recent years, PPM techniques for the prediction of the next activities have matured significantly, mainly thanks to the use of Neural Networks (NNs) as a predictor. 
While their performance is difficult to beat in the general case, there are specific situations where background process knowledge can be helpful.
Such knowledge can be leveraged for improving the quality of predictions for exceptional process executions or when the process changes due to a concept drift. 
In this paper, we present a \textit{Symbolic[Neuro]} system that leverages background knowledge expressed in terms of a procedural process model to offset the under-sampling in the training data. 
More specifically, we make predictions using \textit{NNs with attention mechanism}, an emerging technology in the NN field.
The system has been tested on several real-life logs showing an improvement in the performance of the prediction task.
\end{abstract}

\section{Introduction}
\emph{Predictive Process Monitoring (PPM)}~\cite{DBLP:conf/bpm/Francescomarino18} is a branch of Process Mining (PM). It includes techniques that use event logs 
for predicting how ongoing process executions will continue up to their completion~\cite{Evermann2017,Polatoetal:2016,tax2017predictive}. 
Such predictions can provide valuable input for planning and resource allocation. PPM approaches typically make predictions based on Neural Networks (NNs) and data on: (i) the \emph{sequence of activities} already executed; (ii) the \emph{timestamp} indicating when each activity in the process execution was performed; and (iii) the \emph{values of data attributes} after each occurrence of an activity in the process execution. These techniques are known as techniques for Next Activity Prediction (NAP).

The performance of general NAP techniques has advanced in recent years to a level where prediction accuracy can only be marginally improved~\cite{DBLP:journals/corr/abs-2309-09618}.
What remains a challenge though is 
the correct prediction of trace variants in the test set that are under-sampled in the training set~\cite{DBLP:journals/corr/abs-2309-09618}. This is the case, for example, when the prediction must be performed on process executions that occur in exceptional situations or when the process changes due to a concept drift. Think for instance to the temporary unavailability of a surgery room, which may delay or even rule out the possibility of executing certain activities. 
In such scenarios, NNs would match the under-sampled variants on seen variants during training with the risk of low performance.

In this paper, we address this problem by supporting the prediction task with background knowledge, which is often available in real-life situations.
However, such knowledge has to be considered in a controlled way in order to avoid harming the predictive accuracy of NNs in the general case.
To this extent, we propose to use a \textit{Symbolic[Neuro]} system \cite{DBLP:journals/corr/abs-2105-05330} based on NNs with attention mechanism \cite{vaswani2017attention} reinforced with knowledge-driven modulation. Such knowledge is expressed in terms of a procedural process model (a Petri net, in particular) to offset the log under-sampling when predicting next activities in exceptional process executions.
Our system is based on a beamSearch that guides the construction of the suffix of a trace given its prefix. The key capability of our system in contrast to prior work on beamSearch~\cite{di2017eye} is the use of background knowledge during space exploration. To this end, we combine the predictions of the NN with a compliance score between the predicted suffix and the Petri net model.

The rest of the paper is structured as follows. Section~\ref{sec:background} introduces preliminary concepts that we will use throughout the paper. 
Section~\ref{sec:our_approach} presents the approach we developed for NAP based on NN with attention mechanism. Section~\ref{sec:evaluation} presents the design of our experiments and discusses the results. Section~\ref{sec:related_work} discusses our contributions in the light of prior research on NAP.
Section~\ref{sec:conclusions} concludes the paper with a summary and an outlook on future research. 


\section{Preliminaries}
\label{sec:background}
In this section, we introduce the notions needed to understand the remaining part of the paper. Furthermore, we present a state-of-the-art approach for knowledge-driven NAP that we use as a starting point in this paper.

\subsection{Events, Traces, Logs and Prediction Tasks}
This section provides the general definitions needed for defining the NAP tasks that we address in this paper. 

\begin{definition}[Event] An $event$ is a tuple $(a, c, t)$, where $a \in \mathcal{A}$ is the activity label, $c$ is the trace id, $t$ is the timestamp corresponding to the event.
\end{definition}
We indicate with $\mathbb{E}$ the set that contains all the possible events.
Beyond activity label, trace id, and timestamp, an event may also have additional attributes, such as resources or activity costs. However, for the remainder, a model of the control-flow perspective is sufficient. 

\begin{definition}[Trace] A trace is a non-empty sequence $\sigma = \langle e_1,\ldots,e_{n} \rangle$ of events such that $\forall\ i \in \{1,\ldots,n\}$, $e_i \in \mathbb{E}$ and $\forall \ i,j \in \{1,\ldots,n\}, e_i{.}c = e_j{.}c$, i.e., all events in the sequence refer to the same trace.
\end{definition}
We indicate with $\mathbb{S}$ the set that contains all the possible traces. In the literature on process mining, many business process tasks focus only on the activity labels of a trace. Therefore, it is customary to perform the projection of the activity labels from a trace to a \emph{trace variant}.
\begin{definition}[Trace Variant] A trace variant of a trace $\sigma = \langle e_1, e_2, \ldots, e_n \rangle$ is the sequence of the activity labels in $\sigma$, $\langle e_1.a, e_2.a, \ldots, e_n.a \rangle$.
\end{definition}
We use interchangeably the terms trace and trace variant, and we use the same symbol $\sigma$ for both if there is no risk of ambiguity. In the following, we refer to the activity label $e_i.a$ using the symbol $y_i$, which is the standard way of indicating labels in machine learning. A trace (variant) is, therefore,  $\sigma = \langle y_1, y_2, ... , y_{n} \rangle$.
\begin{definition}[Event log] An \emph{event log} $\mathcal{L} \subseteq \mathbb{S}$ is a set of traces.
\end{definition}

\begin{definition}[Trace length, prefix, suffix]
Given a trace $\sigma = \langle y_1, y_2, ... , y_{n} \rangle$, $l: \sigma \longrightarrow \mathcal{N}$ returns the length of the trace, i.e., the number $n$ of events in $\sigma$. A prefix $\sigma_{\leq k}$ of length $k \in [1, n]$ of  $\sigma$ is the sub-sequence including the first $k$ events of $\sigma$, $\sigma_{\leq k} = \langle y_1, y_2, \ldots, y_{k} \rangle$, with $l(\sigma_{\leq k}) = k$. The suffix corresponding to prefix $\sigma_{\leq k}$, indicated with $\sigma_{>k}$, is the sub-sequence including the remaining events to the end of the trace, $\sigma_{>k} = \langle y_{k+1}, y_2, \ldots, y_n \rangle$, with  $l(\sigma_{>k}) = n-k$.
\end{definition}

\begin{definition}[Prefix Log]
Given an event log $\mathcal{L}$, the \emph{prefix log} $\mathcal{L}^*$ of $\mathcal{L}$ is the event log that contains all prefixes of $\mathcal{L}$, i.e., $\mathcal{L}^* = \{\sigma_{\leq k} : \sigma \in \mathcal{L}, 1 \leq k < l(\sigma) \}$.
\end{definition}
Prefix $\sigma_{\leq k}$ of trace $\sigma$ represents an incomplete ongoing trace in the real world. The goal of the NAP task is to predict the suffix for the ongoing trace $\sigma_{\leq k}$.


\begin{definition}[Next-Activity Prediction (NAP) task]
\label{def:next_act}
Given a prefix $\sigma_{\leq k}$ of a trace $\sigma$, the next activity prediction function $\mathcal{F}: \sigma_{\leq k} \longrightarrow \pi$ returns a probability vector $\pi = [ p(a_{1} | \sigma_{\leq k}), ..., p(a_{i} | \sigma_{\leq k}), ..., p(a_{|\mathcal{A}|}| \sigma_{\leq k}) ]$, for all possible next activities $a_{i} \in \mathcal{A}$, with $i \in [1, |\mathcal{A}|]$. The NAP task for prefix $\sigma_{\leq k}$ returns the activity label $\hat{y}_{k+1}$ given $\sigma_{\leq k}$, where $\hat{y}_{k+1} = \argmax_{a_i \in \mathcal{A}} p(a_i | \sigma_{\leq k})$.
\end{definition}
Here, the probability $p(a_i | \sigma_{\leq k})$ of a symbol given a prefix can be provided by a data-driven predictor such as a Long-Short-Term Memory Neural Network (LSTM)~\cite{HochreiterS97} or a Neural Network with attention mechanism~\cite{vaswani2017attention}, see Section~\ref{sec:related_work}. Such predictors are trained on a prefix log $\mathcal{L}^*$ containing the prefixes $\sigma_{\leq k}$. The target of the prediction is the symbol at position $k+1$ in the whole trace $\sigma$.

\begin{definition}[Suffix Prediction (SP) task]
\label{def:SP}
All traces in an event log $\mathcal{L}$ include an additional completion activity label $\bot$, i.e.,  $\sigma = \langle y_1, y_2, ... , y_{n}, \bot \rangle$. Given a prefix $\sigma_{\leq k}$ of a trace $\sigma$, the suffix prediction function $\mathcal{G}: \sigma_{\leq k} \longrightarrow \overline{\sigma} $ returns a complete trace $\overline{\sigma} = \langle y_1, ..., y_{k}, \hat{y}_{k+1}, ..., \bot \rangle$, where each predicted activity label $\hat{y}$ in suffix $\sigma_{> k}$ is obtained by recurrently applying $\mathcal{F}$ until the completion activity label $\bot$ is observed.
\end{definition}
Predicting a suffix corresponding to the next $n$-steps from a prefix $\sigma_{\leq k}$, i.e., predicting activity label $\hat{y}_{k+n}$, is more difficult when $n$ increases since the uncertainty of this prediction cumulatively increases at each step of the suffix.


\subsection{State-Of-The-Art NAP with Background Knowledge}
\label{sec:SOTA}


Predicting the suffix of a given prefix is a problem that is tackled by state-of-the-art approaches that make use of LSTM-based RNNs~\cite{Evermann2017,tax2017predictive}. 
An approach for taking into account background knowledge when predicting the future path of an ongoing trace was presented in \cite{di2017eye}. Its idea is as follows: 
(i) an LSTM-based RNN is used to get all possible suffixes of the ongoing trace; (ii) the suffixes are ranked according to their likelihood; and (iii) the first suffix that is compliant with the background knowledge is provided as predicted suffix. However, although RNN inference algorithms are not computationally expensive per se, building all the possible predicted suffixes could be inefficient.
Therefore, on top of state-of-the-art LSTM-based techniques, the approach presented in~\cite{di2017eye} uses a \textit{beamSearch} algorithm classically used in statistical sequence-to-sequence predictions in translation tasks~\cite{Tillmann2003WRD778822778827}. The beamSearch is a heuristic algorithm based on graphs that explores the search space of the predicted suffixes by expanding only the most promising branches.

\autoref{alg:protrack} reports the pseudo-code describing the \textsc{SOTA-BS} algorithm proposed in~\cite{di2017eye}. It takes as input a prefix $\sigma_{\leq k}$, the available background knowledge $\cal{BK}$ in the form of Linear Temporal Logic constraints on finite traces ($LTL_f$), and the trained LSTM model, together with three parameters: (i) $bSize$, i.e., the maximum number of next symbols predicted by the LSTM model used to construct the possible predicted suffixes at each iteration; (ii) $maxSize$, i.e., the maximum number of branches that can be explored by the algorithm at the same time; and (ii) $max$, i.e., the maximum number of allowed iterations.

The algorithm builds the suffix given a prefix $\sigma_{\leq k}$ by appending to the prefix a symbol predicted by the LSTM. In addition, it is possible to define a score for the predicted suffix in a recursive way. For the initial prefix $\sigma_{\leq k}$, $score(\sigma_{\leq k})$ is set to 1, whereas $score(\langle\sigma_{\leq m},a\rangle)$, for a suffix with $m \geq k$, is:
\begin{equation}
    score(\langle\sigma_{\leq m},a\rangle) = score(\sigma_{\leq m})\cdot\mathrm{LTSM}(\sigma_{\leq m}, a),
\label{eq:scoreSOTABS}
\end{equation}
where $a \in \mathcal{A}$ and $\mathrm{LTSM}(\sigma_{\leq m}, a)$ is the probability of activity $a$ given $\sigma_{\leq m}$ provided by the LSTM-based RNN.
\begin{algorithm}
	\caption{\textsc{SOTA-BS} algorithm for predicting the suffix of $\sigma_{\leq k}$}
	\label{alg:protrack}
	\begin{algorithmic}[1]
		\Function{\textsc{SOTA-BS}}{$\sigma_{\leq k}$, $\cal{BK}$, $\mathrm{LSTM}$, $bSize$, $maxSize$, $max$}
		\State $h$ = 0
		\State $\mathit{prefixes}$ = \{($\sigma_{\leq k}$, 1)\}\label{lst2:initialization}
		\While {$(h \leq max)$ and (not \textsc{isEmpty}($\mathit{prefixes}$))} \label{lst2:while}
			\State $candidates\_next$ = \textsc{predictNextSymbolsScores}($\mathrm{LSTM}$, $\mathit{prefixes}$, $bSize$) \label{lst2:prediction}
			\State $top\_candidates$ = \textsc{topRank}$(candidates\_next$, $maxSize)$ \label{lst2:top}
           \State \textsc{empty}($\mathit{prefixes}$)\label{lst2:empty}
			\ForAll {$candidate$ in $top\_candidates$} \label{lst2:forall}
					\If {\textsc{last\_symbol}($candidate$) $\neq \bot$}
						\State \textsc{push}($candidate$, $\mathit{prefixes}$)
\Else
					\If {\textsc{check}$(candidate$, $\cal{BK})$} \label{lst2:compliant}
\State \textbf{return} $candidate$

\EndIf
				\EndIf
			\EndFor
			\State $h = h + 1$
		\EndWhile
		\EndFunction
	\end{algorithmic}
\end{algorithm}

Intuitively, the algorithm iterates over a priority queue of prefixes and their scores, which is initialized with the input pair $(\sigma_{\leq k}, 1)$ (line~\ref{lst2:initialization}) and is used for regulating the number of branches to be explored. For each pair prefix-score in $\mathit{prefixes}$, $bSize$ possible next activities and scores are predicted using the LSTM model. The traces are obtained by concatenating the prefix with the corresponding $bSize$ predicted next activities. The scores are obtained by multiplying the score of the prefix with the score of the next activity given the prefix (line 5). In this way, the algorithm generates $\left|\mathit{prefixes}\right|*bSize$ pairs of traces and their scores.

In order to limit the search space, the algorithm ranks the predicted pairs based on their estimated scores and takes only the top $bSize$ ones (line 6). For each of these traces (line 8), if the last symbol predicted is not the end symbol, the trace is added to $\mathit{prefixes}$ (line 10). Otherwise, if the trace is complete, the algorithm checks if it is compliant with the constraints in $\cal{BK}$ (line 12). In this case, the trace (with its score) is returned (line 13).
The algorithm is then iterated until the queue of prefixes and scores is empty or the maximum number of iterations $max$ is reached (line 4).

Note that this algorithm checks the compliance of the predicted trace only when the entire suffix has been predicted. Differently from this approach, we propose a ``modulation'' of the NN output using the background knowledge at each iteration of the SP process. In this way, the background knowledge is not only used as a final (boolean) check on the admissibility of the predicted suffix, but drives the prediction of the suffix while the suffix is being constructed.





\section{Our Proposal}
\label{sec:our_approach}
We now introduce our approach to SP using NNs with attention mechanism whose output signal is modulated via background knowledge.

\subsection{Next-Activity Prediction (NAP) with Attention Mechanism}
\label{sec:NN}
As NN architecture, we use an emerging technology in the NN field based on the attention mechanism, which has shown promising results in text generation~\cite{vaswani2017attention}. In general, the attention mechanism goes beyond the classical dense layers, convolutional layers and LSTM cells and introduces an entirely new general computing mechanism to model relationships in the data.

The architecture we adopt is based on a transformer encoder that is composed of a stack of 6 identical layers. Each layer has two sub-layers where the first is a multi-head self-attention mechanism, and the second is a simple fully connected feed-forward network. A residual connection around each of the two sub-layers is used, followed by a normalization layer. The output of these layers have a dimension of 64 and the number of heads in the multi-head self-attention mechanism is 8. More details can be found in~\cite{vaswani2017attention}. This encoder is followed by a global max pooling (1D) layer with a dropout layer. Finally, a fully connected layer (with a softmax activation) predicts the probabilities of the activity labels.

This architecture is trained on a prefix log to predict the next activity label given a prefix as input. The prefix log is, therefore, encoded into numeric vectors by using the positional encoding described in the next section.


\subsection{Trace Encoding}
NN works with numeric inputs, therefore we need to encode traces and corresponding 
activity labels as numeric vectors. 
Since an activity label is a categorical attribute, the activities in traces can be encoded as binary values using a one-hot encoding. This means that for each activity label $y_k$ in trace $\sigma = \langle y_1,...,y_{n} \rangle$, a numeric vector $x_k = [c_{k,1}, ..., c_{k,|\mathcal{A}|}]$ is built, where $c_{k,i}$, for $i = 1,...,|\mathcal{A}|$, are binary values defined as follows:

\begin{equation*}
c_{k, i}(y_k) =
\begin{cases}
1 & \text{if } y_k = a_i \in \mathcal{A},\\
0 & \text{otherwise}
\end{cases}
\end{equation*}

The NAP task requires a prefix as input and produces the next activity label as output. Therefore, we use prefixes to train the NN (i.e., a prefix log). 
In particular, for each encoded trace represented by the numeric vector $v = \langle x_1, ..., x_{n} \rangle$, we construct $n-1$ prefixes  $v_{k} = \langle x_1, ..., x_{k} \rangle$, for all $k \in [1,n-1]$. Since vectors $v_{k}$ have different lengths  $k \in [1, l^{Max}-1]$ (where $l^{Max}$ is the maximum trace length in the event log), and, to train the NN, we need to provide a fixed-size matrix as input, we apply zero-padding to all prefixes of length $k$ with $k < l^{Max}-1$.

\subsection{Knowledge-Driven Modulation of Neural Networks with Attention Mechanism for NAP}

As described in \autoref{sec:SOTA}, the \textsc{SOTA-BS} algorithm discards some predicted suffixes because they are not compliant with the background knowledge expressed in the form of $LTL_f$ constraints. This check is ``crisp'', i.e., it returns a boolean value, true or false. The compliant trace with highest score is therefore predicted. However, in some cases, this boolean check may be too strict since it filters out some predicted suffixes that can represent valid future process executions even if they do not satisfy all the $LTL_f$ constraints. When we represent the background knowledge in the form of procedural models, the above mechanism is impossible to be applied since a perfect compliance of the predicted prefix with the background knowledge is, in this case, hard to obtain.

\begin{algorithm}[t]
	\caption{Our proposed \textsc{KB-Modulation} algorithm for predicting the suffix of $\sigma_{\leq k}$}
	\label{alg:KBModulation}
	\begin{algorithmic}[1]
		\Function{\textsc{KB-Modulation}}{$\sigma_{\leq k}$, $\mathcal{BK}$, $\mathrm{AttentionNN}$, $bSize$,$max$, $w$}
		\State $h$ = 0
		\State $prefixes$ = \{$(\sigma_{\leq k}, 1)$\}\label{lst3:initialization}
		\While {$(h \leq max)$ and (not \textsc{isEmpty}($prefixes$))} \label{lst3:while}
			\State $candidates\_next$ = \textsc{predictNextSymbolsScores}($\mathrm{AttentionNN}$, $prefixes$, $bSize$)
	\State $score\_next = \textsc{ExtractScores}(candidates\_next)$
   \State $compliance\_next$ = \textsc{Compliance}($\mathcal{BK}$, $candidates\_next$)
            \State $total\_score\_next$ = $score\_next^{1-w}\cdot compliance\_next^w$
			\State $top\_candidates$ = \textsc{topRank}$(candidates\_next$, $total\_score\_next$, $bSize)$ \label{lst2:topKBMod}

           \State \textsc{empty}($prefixes$)\label{lst3:empty}
           \State $rank$ = 0
			\ForAll {$candidate$ in $top\_candidates$} \label{lst3:forall}
                \If {\textsc{last\_symbol}($candidate$) $\neq$ $\bot$}
                    \State \textsc{push}($candidate$, $prefixes$)
                \Else
                   \If {$rank == 0$}
                    \State \textbf{return} $candidate$

                    \EndIf
				\EndIf
                \State $rank = rank + 1$
			\EndFor
			\State $h = h + 1$
		\EndWhile
		\EndFunction
	\end{algorithmic}
\end{algorithm}


To solve this problem, we propose to measure the conformance of the predicted trace with the background knowledge by using metrics with values in $[0,1]$ where 0 indicates no compliance at all and 1 indicates full compliance. Moreover, to improve the impact of the background knowledge $\mathcal{BK}$, we perform conformance checking during the beamSearch to modulate the predictions of the NN while the suffix is still being constructed. This is different from the \textsc{SOTA-BS} algorithm, where the check is performed only when the suffix has been entirely predicted. We, therefore, modify the score defined in Eq. \eqref{eq:scoreSOTABS} as:

\begin{equation}
score(\langle\sigma_{\leq m},a\rangle) = score(\sigma_{\leq m})\cdot\mathrm{NN}(\sigma_{\leq m}, a)^{1-w}\cdot\mathcal{C}(\langle \sigma_{\leq m},a\rangle, \mathcal{BK})^w,
\end{equation}
where $\mathrm{NN}(\sigma_{\leq m}, a)$ is the probability of activity $a$ given the prefix $\sigma_{\leq m}$ returned by the trained $\mathrm{NN}$ and function $\mathcal{C}(\langle \sigma_{\leq m},a\rangle, \mathcal{BK})$ is a compliance function quantifying the distance between the extended trace with the new symbol $\langle \sigma_{\leq m},a\rangle$ and the background knowledge $\mathcal{BK}$ having values in $[0,1]$. This function can be a fuzzy $LTL_f$ checker~\cite{frigeri2014fuzzy} (if we use declarative models), or a fitness function used in conformance checking algorithms~\cite{DBLP:journals/is/RozinatA08} (if we use procedural models). The $\mathcal{C}$ function modulates the output signal of the $\mathrm{NN}$ with the background knowledge. The parameter $w \in [0, 1]$ is a hyperparameter that sets the power of such a modulation, i.e., it gives more weight to the NN or to the background knowledge.

\autoref{alg:KBModulation} shows our proposal.
Differently from the \textsc{SOTA-BS} algorithm, \textsc{KB-Modulation} modulates the score of each candidate suffix with the fitness between the prefix and the background knowledge available (lines 6 to 9). The fitness value in $[0, 1]$ is, therefore, used while the predicted suffix is still being constructed by injecting the knowledge at every step of the beamSearch. In \textsc{KB-Modulation}, when a trace ends, i.e., when the $\bot$ symbol has been predicted, it is returned only if it has the highest score among all candidates (line 16). Otherwise, the beamSearch is continued until a suffix ending with $\bot$ is predicted with the highest score.

To instantiate this generic algorithmic design, we use a \textit{Neural Network with attention mechanism}, background knowledge $\mathcal{BK}$ in the form of a \textit{Petri net model}, and the compliance function $\mathcal{C}$ provided in~\cite{berti2019reviving} using token-based replay. We use token-based replay instead of alignments to measure the fitness of the predicted trace with the background knowledge for two reasons: (1) we cannot use alignments because we need to use this function while the suffix is being constructed, whereas the fitness function based on alignments requires the input trace to be a complete trace; and (2) as shown in~\cite{berti2019reviving}, computing the fitness using token-based replay is generally faster.


\section{Evaluation}
\label{sec:evaluation}
For our experiments, we used a synthetic log
and 5 real-life logs: Sepsis, RoadTraffic, Helpdesk, BPIC2013 (I), BPIC2013 (CP), BPIC2012.\footnote{The event logs are available at: \\ Sepsis: {\scriptsize \url{https://doi.org/10.4121/uuid:915d2bfb-7e84-49ad-a286-dc35f063a460}}, \\ RoadTraffic: {\scriptsize \url{https://doi.org/10.4121/uuid:270fd440-1057-4fb9-89a9-b699b47990f5}}, \\ Helpdesk: {\scriptsize  \url{https://doi.org/10.4121/uuid:0c60edf1-6f83-4e75-9367-4c63b3e9d5bb}},\\ BPIC2013 (I): {\scriptsize \url{https://doi.org/10.4121/uuid:500573e6-accc-4b0c-9576-aa5468b10cee}}, \\ BPIC2013 (CP): {\scriptsize \url{https://doi.org/10.4121/uuid:c2c3b154-ab26-4b31-a0e8-8f2350ddac11}},\\ BPIC2012: {\scriptsize \url{https://doi.org/10.4121/uuid:3926db30-f712-4394-aebc-75976070e91f}}.} The code we use for the experiments is available for reproducibility at \url{https://github.com/jonghyeonk/KB-Modulation}. As explained in \cite{DBLP:journals/corr/abs-2310-18979}, real-life
datasets are valuable for capturing the variability and complexity of a real-world problem, whereas synthetic datasets can be used once relevant input data features
are understood and when empirical data does not exist, is insufficient, imprecise, or too large to store or share.
For testing our \textsc{KB-Modulation} algorithm, we generated a synthetic log to reproduce the ideal situation where a (procedural) background knowledge is available and is constructed to perfectly characterize the exceptional process executions that we want to predict in the test set (this knowledge is not available in the case of the real-life datasets).

\subsection{Synthetic Log Dataset}
\label{sec:synthetic}

The synthetic log was generated to contain variants that are infrequent in the training set, but are frequent in the test set. In this way, we can test the performance of the algorithm in the prediction of exceptional process executions. In particular, the synthetic log was generated using the following procedure:
\begin{itemize}
    \item We defined a basic BPMN model containing 9 activity labels and two main exclusive branches. Each branch contains activities in sequence or in parallel. Each branch contains also an exclusive choice between a sequence of special activities, i.e., \texttt{Unexpected} followed by \texttt{Repairing}, and the normal execution. This means that, in some cases, an unexpected event happens and it must be followed by a repairing event. The complete model can be found in our repository at \url{https://github.com/jonghyeonk/KB-Modulation}.
    \item The synthetic dataset has been generated by using the PLG tool~\cite{BurattinS10}: 1000 traces have been generated for the training set in a way that it contains 800 traces corresponding to normal executions and only 200 traces (20\% of the entire set) containing the sequence \texttt{Unexpected}-\texttt{Repairing}.
    \item The test set contains the 200 traces of the training set corresponding to the unexpected executions.
\end{itemize}
A classical beamSearch algorithm for the SP task will predict the most frequent variants in the training set (those with a normal execution) starting from a prefix of the test set. Therefore, using a test set built as described, this would result in poor performance as it contains only traces with unexpected executions. This is overturned by using our proposed \textsc{KB-Modulation} algorithm.

\subsection{Real-Life Log Datasets}
\label{sec:real}
\begin{figure}[t!]
\centerline{\includegraphics[width=0.75\linewidth]{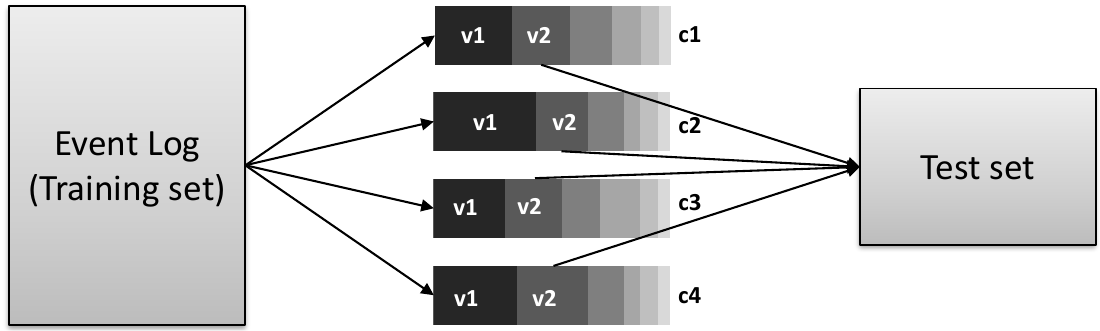}}
\caption{Procedure to build the Real-Life datasets.}
\label{fig:ExperimentalSetting}
\end{figure}

For the real-life logs, we split each log in a training and in a test set (for each prefix length $k$ we split the log in a different way) and then we discover the Petri net model representing the background knowledge from the test set. Note that, for each prefix length $k$, the training set is the entire (prefix) log. The test set is, instead, used as is to discover the Petri net model, while, for testing purposes, we use only prefixes of length $k$ and consider the complete traces as ground truth.

The rationale behind the construction of the test set, for each prefix length $k$, is that the set is constructed using traces that are infrequent in the training set and have the same prefix (of length $k$) as high-frequent traces but different suffix. In this way, since the test set includes only less frequent cases, the trained NN will predict, for the given prefix length, the suffix provided by the high-frequent traces (that were available in the training set). Therefore, the information contained in the background knowledge (built from the test set and, therefore, excluding the high-frequent traces) will help in finding the correct suffix.

In particular, for each prefix $k$, we follow the procedure shown in \autoref{fig:ExperimentalSetting}. We cluster the traces in the log by including in the same cluster all traces having the same prefix of length $k$. Then, we group, in each cluster, the traces per variant and we sort the variants according to the number of traces included in each variant. We then build the test set by taking from each cluster the variant that is the second in the list of the variants ($v2$) ordered by frequency. These traces will have the same prefix as traces in the most frequent variant ($v1$), but a suffix that is less frequent in the training set. In this way, if we discover from the test set the Petri net model that we use as background knowledge, for prefixes of length $k$, the NN tends to favour the suffix provided by traces in the most frequent variant in the training set having that prefix ($v1$), but the background knowledge will correct this prediction.

\subsection{Experimental Settings}


We run our experiments for prefix lengths ranging from 3 to 7, 
and we implemented a three-folds replication in the training for generalization. We then run the \textsc{KB-Modulation} algorithm by varying the input $w \in [0, 0.1, ..., 0.95]$, with $bSize = 3$, and with $max$ corresponding to the maximum trace length in each log. We considered as baseline method a basic beamSearch algorithm without $\mathcal{BK}$ using the same multi-headed attention network we used to run the \textsc{KB-Modulation} algorithm.
The Petri net used for representing the $\mathcal{BK}$ was discovered using the Split Miner~\cite{augusto2019split}. In particular, we used the discovery tool in \cite{augusto2018automated} supporting automatic hyperparameter optimization based on the input log.


Regarding the evaluation metrics, the tested technique predicts a suffix ($\hat{\sigma}_{>k}$) given a prefix ($\sigma_{\leq k}$). Therefore, for the evaluation, we measured the similarity between the predicted suffix $\hat{\sigma}_{> k}$ and the ground truth suffix $\sigma_{>k}$ using the Damerau-Levenshtein similarity~\cite{damerau1964technique} for all prefixes in the test set, and then we averaged them. In addition, since we construct different test sets for different prefix lengths, the similarity scores over different prefix lengths were again averaged using a \textit{micro average}, i.e., by weighting the contribution of each value based on the size of the corresponding test set.

\subsection{Results}

\begin{figure}[t]
\centering
\includegraphics[width=0.85\linewidth]{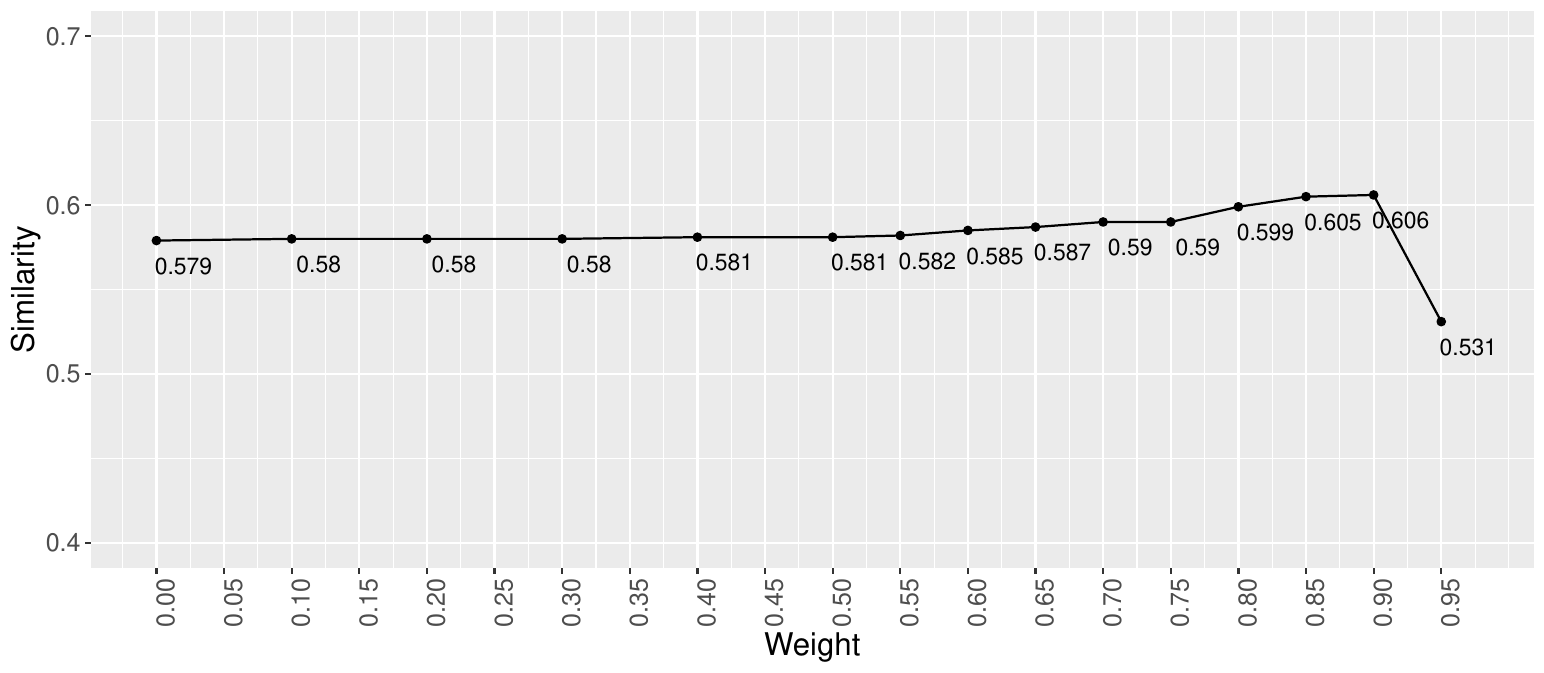}
\vspace{-0.4cm}
\captionof{figure}{Trend of similarity (computed as micro-average over the different prefixes) for different weight values $w$ (in the case of the Sepsis dataset).}
\label{fig:trend}
\end{figure}

In \autoref{fig:trend}, we show the trend of the (micro-average) similarity between predicted suffixes and ground truth suffixes in the case of the Sepsis dataset. The trend is similar for all the datasets, i.e., when the weight of the fitness with respect to the background knowledge increases, the similarity (and the performance of the prediction task) improves. This is related to the fact that all the datasets are built in a way that the test set is fully compliant with the Petri net model representing the background knowledge. Note that when $w>0.9$ the similarity goes quickly down since the background knowledge alone without the contribution of the NN score is not able to discriminate between variants with different frequencies. This phenomenon is related to the fact that the experiments we conducted were designed to predict exceptional but still reasonably frequent process executions. As we will see in the following discussion, the exceptional executions need to have a certain frequency in the training set, otherwise the fitness score does not manage to compensate the NN score if this is close to zero. This makes the proposed approach particularly useful in the presence of concept drifts in the process, i.e., when, due to seasonal changes in the environment where the process is executed, variants that were previously less frequent become frequent.

In \autoref{tab:table}, we show (1) the similarity for $w=0$, i.e., for the case in which we do not use the background knowledge (baseline), and (2) the best similarity obtained by varying $w$ in $[0.1, ..., 0.95]$, for the case in which we use it (BK).
We show the similarity values per prefix and aggregated using the micro average.

\begin{table}[t]
\centering
\captionof{table}{Suffix similarity with the baseline and the knowledge-driven approach.}
\label{tab:table}
\resizebox{1.0\linewidth}{!}{
\begin{tabular}{L{1.8cm}||C{1.7cm}|C{1.3cm}|C{1.3cm}|C{1.1cm}||C{2.2cm}|C{1.3cm}|C{1.3cm}|C{1.1cm}}
\hline
\multirow{2}{*}{Data} & \multicolumn{1}{c|}{\multirow{2}{*}{Prefix size}} & \multicolumn{2}{c|}{Suffix Similarity} & \multirow{2}{*}{Diff} & \multirow{2}{*}{Prefix size} & \multicolumn{2}{c|}{Suffix Similarity} & \multirow{2}{*}{Diff} \\ \cline{3-4} \cline{7-8}
 & \multicolumn{1}{l|}{} & \multicolumn{1}{c|}{baseline} & BK &  &  & \multicolumn{1}{c|}{baseline} & BK &  \\ \hline
 & 3 & 0.577 & 0.811 & \textbf{0.234} & 6 & 0.542 & 0.891 & \textbf{0.349} \\
Synthetic & 4 & 0.548 & 0.796 & \textbf{0.248} & 7 & 0.502 & 0.840 & \textbf{0.339} \\
 & 5 & 0.534 & 0.767 &\textbf{0.232} & micro average & 0.540 & 0.814 & \textbf{0.274} \\ \hline
 & 3 & 0.606 & 0.682 & \textbf{0.076} & 6 & 0.577 & 0.600 & 0.023 \\
Sepsis & 4 & 0.643 & 0.654 & 0.011 & 7 & 0.517 & 0.573 & \textbf{0.057} \\
 & 5 & 0.619 & 0.628 & 0.009 & micro average & 0.579 & 0.609 & 0.030 \\ \hline
 & 3 & 0.496 & 0.497 & \textit{0.001} & 6 & 0.299 & 0.318 & 0.020 \\
RoadTraffic & 4 & \underline{0.042} & 0.043 & \textit{0.001} & 7 & 0.618 & 0.777 & \textbf{0.159} \\
 & 5 & \underline{0.047} & 0.049 & \textit{0.002} & micro average & 0.229 & 0.233 & \textit{0.004} \\ \hline
 & 3 & 0.596 & 0.774 & \textbf{0.177} & 6 & 0.447 & 0.447 & \textit{0.000} \\
Helpdesk & 4 & 0.593 & 0.598 & \textit{0.005} & 7 & 0.397 & 0.404 & 0.007 \\
 & 5 & 0.583 & 0.597 & 0.014 & micro average & 0.585 & 0.686 & 0.101 \\ \hline
 & 3 & 0.474 & 0.503 & 0.029 & 6 & 0.550 & 0.660 & \textbf{0.110} \\
BPIC2013 & 4 & 0.650 & 0.662 & 0.013 & 7 & 0.515 & 0.521 & 0.006 \\
(I) & 5 & 0.618 & 0.632 & 0.014 & micro average & 0.561 & 0.576 & 0.015 \\ \hline
 & 3 & 0.732 & 0.732 & \textit{0.000} & 6 & 0.527 & 0.527 & \textit{0.000} \\
BPIC2013 & 4 & 0.399 & 0.417 & 0.018 & 7 & 0.435 & 0.438 & \textit{0.003} \\
(CP) & 5 & 0.411 & 0.411 & \textit{0.000} & micro average & 0.557 & 0.558 & \textit{0.001} \\ \hline
 & 3 & 0.177 & 0.731 & \textbf{0.554} & 6 & \underline{0.091} & 0.132 & 0.041 \\
BPIC2012 & 4 & 0.143 & 0.665 & \textbf{0.522} & 7 & \underline{0.063} & 0.107 & 0.044 \\
 & 5 & \underline{0.092} & 0.652 & \textbf{0.561} & micro average & 0.119 & 0.481 & \textbf{0.362} \\ \hline
\end{tabular}
}
\vskip -0.1 in
\end{table}

The results show that for the synthetic log that was built using a scenario where the background knowledge can affect significantly the quality of the predictions, the similarity between the predicted suffixes and the ground truth is much higher when using the background knowledge with respect to the case in which the background knowledge is not used. In this case, we have an average improvement of 0.274 in terms of similarity.

For what concerns the real-life logs, we can see different behaviors. In some cases (identified with italic font), there is no significant difference ($\mathit{diff} \leq 0.005$) between the baseline and the knowledge-driven predictions. This is due to the fact that, in those cases, the second best variants (used to build the test set) have very low frequency with respect to the most frequent variants in each cluster. This implies that, given a prefix, the difference between the probability score given from the NN to the suffix corresponding to the most frequent variant in the training set and the one given to the suffix corresponding to the second best variant is so significant that, even if the fitness score of the latter is higher than the fitness score of the former, the higher fitness score does not manage to compensate the difference in terms of probability (as also mentioned at the beginning of this section). For this reason the background knowledge is less effective and the improvement is less evident. Note that since the test set is composed of multiple second best variants, this phenomenon is more evident when there are several clusters in which the second best variant has very low frequency with respect to the most frequent variant in the same cluster. For the cases in bold, instead, the background knowledge is effective ($\mathit{diff} \geq 0.05$), because the scenario explained above does not occur and the higher fitness is able to correct the wrong predictions of the NN.

For some prefixes (indicated with underlined font), the similarity score is close to 0 ($< 0.1$). This is due to the fact that, as explained above, the similarity score is computed only on the suffix (the predicted part). For this reason, in the cases in which the suffix is short (in the extreme case it contains only one activity label), a mistake in the predicted labels dramatically affects the similarity that becomes close to zero. This happens for the RoadTraffic dataset. In other cases, the performance of the NN is rather poor for the presence of loops (this is a well-known problem in the literature \cite{tax2017predictive}) with activity labels repeated several times (this is observed for the BPIC2012).


\section{Related Work}
\label{sec:related_work}
Following the work in~\cite{DBLP:conf/bpm/Francescomarino18}, the perspectives of PPM can be classified into the four types, i.e., (i) the activity perspective to predict a suffix of incomplete cases, (ii) the time perspective to predict remaining execution time of incomplete cases, (iii) the risk perspective to provide prediction and recommendation to reduce risks, and (iv) the outcome perspective to predict a business objective. Our approach falls into the first perspective (we focus on the NAP and SP tasks). 

The approaches for NAP existing in the literature use decision trees based on sequential pattern mining~\cite{ceci2014completion}, probabilistic models based on Markov chains \cite{le2012hybrid,DBLP:conf/emisa/BeckerBDM14,lakshmanan2015markov,unuvar2016leveraging}, Probabilistic
Finite Automata~\cite{breuker2016comprehensible}, Deep Neural Networks integrated with process discovery~\cite{theis2019decay}, Convolutional Neural Networks~\cite{pasquadibisceglie2019using}, and Recurrent Neural Networks with Long Short-Term Memory \cite{Evermann2017,tax2017predictive,ketyko2022averages,khan2018memory,lin2019mm,taymouri2021deep}. Most of these works have been extended to support the SP task. For example, Lakshmanan et al.~\cite{lakshmanan2015markov} propose an approach able to predict the suffix of a running case and its outcome. They also addressed how to fit a probabilistic model containing loops and parallelism. Khan et al.~\cite{khan2018memory} adapt the Differential Neural Computer~\cite{graves2016hybrid}, which is a Memory–Augmented Neural Network (MANN) to equip an external memory for LSTM. Lin et al.~\cite{lin2019mm} propose an SP framework based on two LSTM networks used as encoder and decoder. Then, by adding modulators between the two LSTM layers and providing weights to events and their attributes, they show improved performance with respect to state-of-the-art approaches. Taymouri et al.~\cite{taymouri2021deep} propose a beamSearch algorithm with LSTM-based RNNs equipped with adversarial loss functions for adversarial LSTM training.

The approaches just described support the SP task without using background knowledge. Di Francescomarino et al.~\cite{di2017eye}, instead, propose a beamSearch algorithm with LSTM-based RNNs, where the background knowledge is used in the form of $LTL_f$ rules. The background knowledge provides an additional check that ensures that the predicted suffix fits the specified $LTL_f$ rules. In our algorithm, instead of checking the compliance of the predicted trace with the background knowledge using a boolean check when the suffix has been entirely predicted, the construction of the suffix is knowledge-driven via the integration of a fuzzy fitness function to select the best predicted activity labels at each iteration.

\section{Conclusion}\label{sec:conclusions}
In this paper, we showed how to build a \textit{Symbolic[Neuro]} system based on \textit{NNs with attention mechanism} for NAP. This approach represents an advancement with respect to the state-of-the-art for several reasons. Most of the state-of-the-art approaches for NAP do not consider the possibility of having a background knowledge that under exceptional circumstances (e.g., in the case of a concept drift in the process) can correct the prediction of the NN. The only existing approach that takes into consideration background knowledge \cite{di2017eye} uses a very basic strategy where the compliance with the background knowledge is performed \textit{after the prediction phase} and with a \textit{boolean check} that filters out all the predictions that are not fully compliant with it. Our proposal, instead, uses the background knowledge while the predicted suffix is being constructed by using a (fuzzy) fitness score in combination to the score provided by the NN to guide the construction of the suffix. The above strategy allows us to represent the background knowledge in the from of a procedural model (a Petri net model), which would be impossible to be used when filtering the predicted traces with a ``crisp'' checker. Finally, for the first time we implement the NAP and SP tasks using NNs with attention mechanism.

As future work, we would like to implement \textsc{KB-Modulation} using a fuzzy $LTL_f$ checker~\cite{frigeri2014fuzzy} (making it directly comparable with the \textsc{SOTA-BS} algorithm) and use the four-valued semantics presented in \cite{DBLP:conf/bpm/MaggiMWA11} in the compliance function. We will also enrich our proposal by considering not only the control-flow perspective, but also the data (and time) perspective both in the trace encoding and in the background knowledge representation.
Finally, we would like to define a systematic method to choose the optimal value of weight $w$ to tune the influence of the background knowledge on the predicted suffixes based on the frequency of the corresponding process executions in the training set.  

\bibliographystyle{splncs04}
\bibliography{main}
\end{document}